# Crack Semantic Segmentation using the U-Net with Full Attention Strategy

Fangzheng Lin[1*], Jiesheng Yang[1], Jiangpeng Shu[2], and Raimar J. Scherer[1]
[1] Institute of Construction Informatics, Technische Universität Dresden, Dresden, Germany
[2] College of Civil Engineering and Architecture, Zhejiang University, Hangzhou, China

**ABSTRACT**

Structures suffer from the emergence of cracks, therefore, crack detection is always an issue with much concern in structural health monitoring. Along with the rapid progress of deep learning technology, image semantic segmentation, an active research field, offers another solution, which is more effective and intelligent, to crack detection Through numerous artificial neural networks have been developed to address the preceding issue, corresponding explorations are never stopped improving the quality of crack detection. This paper presents a novel artificial neural network architecture named Full Attention U-net for image semantic segmentation. The proposed architecture leverages the U-net as the backbone and adopts the Full Attention Strategy, which is a synthesis of the attention mechanism and the outputs from each encoding layer in skip connection. Subject to the hardware in training, the experiments are composed of verification and validation. In verification, 4 networks including U-net, Attention U-net, Advanced Attention U-net, and Full Attention U-net are tested through cell images for a competitive study. With respect to mean intersection-over-unions and clarity of edge identification, the Full Attention U-net performs best in verification, and is hence applied for crack semantic segmentation in validation to demonstrate its effectiveness.

**INTRODUCTION**

Convolutional Neural Networks (CNNs) are driving advances in recognition. The corresponding technologies are absorbed relatively quickly in engineering practice, such as crack detection, which is a vital concern in structural health monitoring. Silva and Lucena [1] trained a VGG-16 [2] based model with limited data to realize concrete cracks detection. The YOLO v2 net [3] is applied to mark cracks from Photos in bounding boxes, as [4] states. From object identification to bounding box detection, the quality of crack detection results is more and more highly expected with the development of CNNs. As an advanced requirement in crack detection, adopting CNNs to segment cracks in images is nowadays one of the most popular and challenging topics.

The literature review indicates that Fully Convolutional Networks (FCNs) [5] and U-net [6] are two mainstream architectures utilized to manage image semantic segmentation. FCN is a deep learning network proposed for generic segmentation tasks. It is named by the original proposal, where fully connected layers in a classifier are replaced by convolution layers. In the very first paper [5], AlexNet [7], the VGG net [2], and GoogLeNet [8] are transformed and adopted for validation using common datasets such as PASCAL VOC, NYUDv2 and SIFT Flow. With specific focusing on cracks, the feasibility of FCNs also has been convinced. Dung and Anh [9] applied FCNs through three pre-trained different network architectures to realize identification and density calculation of concrete cracks. Considering a more intensive practice scenario in metro tunnels, Huang et. al. [10] developed a two-stream algorithm via FCN to detect the regions of cracks and leakages separately. Generally, semantic segmentation using FCN should be completed in two steps: training a CNN model, and then transforming the trained model with a FCN. Hence, a large amount of training dataset is usually required, e.g. 40,000 227×227 pixel images in [9], 188,704 and 110,466 images of 500×500 pixel resolution respectively for cracks and leakages in [10].

The U-net architecture, which is initially implemented in biomedical image segmentation in 2015 [6], offers another kind of solution. Obeying the encoder-decoder philosophy involved by FCNs as well,

U-net emphasizes the structural symmetry of networks, both pooling part (encoder) and upsampling part (decoder) have 4 layers. Ever since it was brought to the field of crack detection, U-net has demonstrated its efforts. The original U-net architecture was employed in [11]. In that case, the Adam algorithm was used in optimization of training procedure, k-Fold cross validation is used for evaluation. It is to be noticed that Liu et. al. [11] took merely 84 images of 512×512 pixels resolution.

Certainly, enrichment of the U-net becomes a research tendency. For instance, attention gating between encoder and decoding section [12], and residual connections in each encoder and decoder block [12, 13] were added in the proposed U-net networks to improve the results of semantic segmentation. 117 labeled images of size 480×320 pixels and 38 images of mixed pixels were fed for training in [12]. In comparison with conventional FCNs, U-net demands much fewer data in training and shows comparable validation results. As another popular tendency, U-net based network reconstruction was explored through combination [14, 15], parallel execution [16], coupling [17, 18], etc. An advanced neural network named as U-net++ is proposed [19]. U-net++ targets on the inherent shortcomings of U-net namely inefficiency in multi-task learning and unnecessarily restrictive skip connection in networks, realizes an essential update in terms of network architecture.

The state-of-the-art approaches show the further enrichment potential and feasibility of the U-net architecture. With respect to that, we propose two novel architectures inspired by the U-net network. Full attention strategy is invented and brought into the novel architectures to yield a more optimal accuracy in semantic segmentation. Due to limited computing ability, we designed the experiment in two parts, verification and validation. In verification, a set of open source cell images is adopted to investigate the performance of the original U-net, the Attention U-net, and two proposed novel networks. The best network was then applied in validation for crack image segmentation.

This research work contributes to Project 1 in the 1st international project competition (IPC-SHM 2020). Section MODELLING FOR SEGMENTATION clarifies the basic principles of the U-net and attention mechanism, elaborates the idea of proposed novel networks. The training configuration and process are depicted in section MODEL TRAINING. In section EXPERIMENTS, the prediction results of verification and validation are demonstrated and interpreted, the segmentation quality of 4 networks are discussed and evaluated. Finally, it goes to the discussion and conclusions of the entire research work and forecasts the possible directions of future work.

**ARCHITECTURES FOR SEGMENTATION**

U-net architecture (see Fig. 1) is chosen to be utilized and further developed to address image semantic segmentation. In this section, two classic networks, the original U-net and Attention U-net, are introduced. The Full Attention Strategy is elaborated and two novel U-net based architectures are conceived.

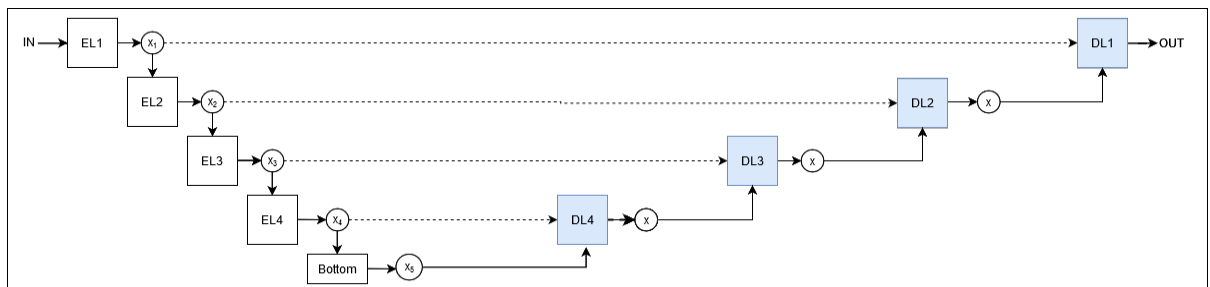

Fig. 1. Architecture of the original U-net

*U-Net Architecture*

The squares in Fig. 1 indicate the down- and upsampling (i.e. encoding and decoding) layers of the original U-net. $EL$ denotes the encoding layer, $DL$ the decoding layer. $x_{i=1,2,3,4,5}$ and $x$ represent the output of respectively the encoding and decoding layers. In each layer of downsampling, input data should be processed by an operation loop twice, which encompasses a convolutional layer, a batch normalization layer and the ReLU activation function. The first loop is utilized to transform the size of input to a longer tensor with less resolution, the tensor is not to be extended in the second loop. Max pooling, a widely-speared pooling strategy, is chosen to reduce the dimensionality of input furthermore.

The upsampling layers on the right side of the proposed architecture is basically symmetric to the left side. Each layer, marked by blue in Fig. 1, contains concatenation operation, two loops composed of the same functions as those in downsampling, and a deconvolution layer. In the bottom layer, there are merely upsampling loops and a deconvolution layer but no concatenation. The deconvolution layer in the upset upsampling layer is replaced by a convolutional operation and a Softmax activation function. U-net has shown comparable segmentation ability and prominent advantages among the CNN architectures in semantic segmentation [9, 10]. However, the state-of-the-art papers state, there is still potential that has not been exploited sufficiently.

*Attention Mechanism*

The basic principle behind U-net in image segmentation lays in feature extraction. During model training, noise in images should be filtered out as much as possible, meanwhile, the features of interest are supposed to be focused and extracted. Hence, the attention mechanism (see Fig. 2), which can be incorporated into U-net, is an effective approach to highlight only the relevant regions of an image during training. It reduces the computational resources wasted on irrelevant activations, providing the network with better generalization power [20]. The attention mechanism in U-net belongs to soft attention, with which several regions in a picture can be highlighted. Soft attention works by weighting different parts of the image. Areas of high relevance are multiplied with a larger weight and areas of low relevance are tagged with smaller weights.

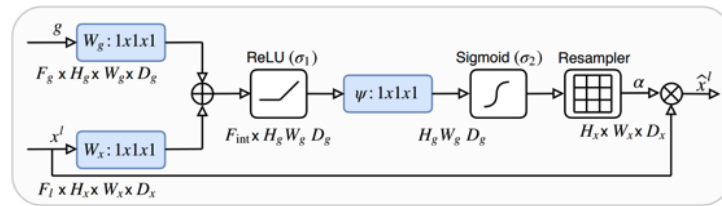

Fig. 2. Attention gate schematic [20]

Attention mechanism is implemented through an attention gate (AG) to capture essential features. AGs represented by yellow triangles in Fig. 3 are added before the concatenation operation in each decoding of the Attention U-net. The output from a corresponding downsampling layer over skip connection is represented as $g$, that from the last upsampling layer as $x^l$, both serve as input of an Attention Gate. After convolutional operation and processing with a sigmoid activation function, the partial highlighted image is subsequently produced and goes to the next step, concatenation. The picture sizes of inputs and outputs of an AG should be unchanged and match the size requirement of the next operation.

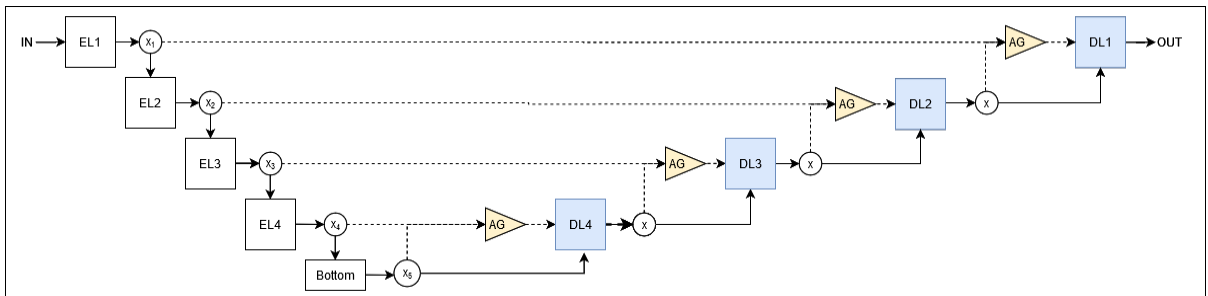

Fig. 3. Architecture of Attention U-net

*Utilization of Skip Connections*

Huang et al. proposed an architecture named U-net 3+, where the outputs from encoding layers are adopted by each decoding layer, and the output of the bottom layer is also applied by related decoding layers. Inspired by the research work, we suggested an Advanced Attention U-net (see Fig. 4), where the outputs from encoding layer 1, 2 and 3 are maxpoolled and convoluted to satisfied the required depth of each decoding layer. Before the outputs go to the attention gate, the original and operated outputs are to be concatenated, the concatenated operation is depicted with red blocks in Fig. 4. The performance of the Advanced Attention U-net will be explored in the experiments and compared with other networks.

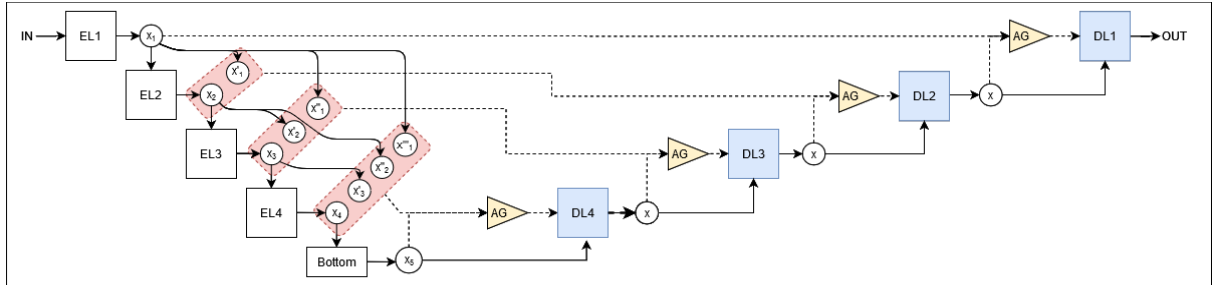

Fig. 4. Architecture of Advanced Attention U-net

*Full Attention Strategy*

Advanced U-net takes advantage of skip connection in U-net, and applies more information to facilitate training. Theoretically, more information in training means that the more features are delivered to each decoding layer. Nevertheless, the information consists of not only features of interest, but also an amount of noise. As a consequence, both features and interference objects in noise are more intensely focused, and some existing noise is able to be retained instead of filtered off. Since the noise proportion declines through AGs, we put an AG after each output. We name the combination of utilization of outputs from encoding layers and Attention Gates Full Attention Strategy as Full Attention Strategy. Full Attention U-net in Fig. 5 is derived from Advanced Attention U-net, and adopts Full Attention Strategy

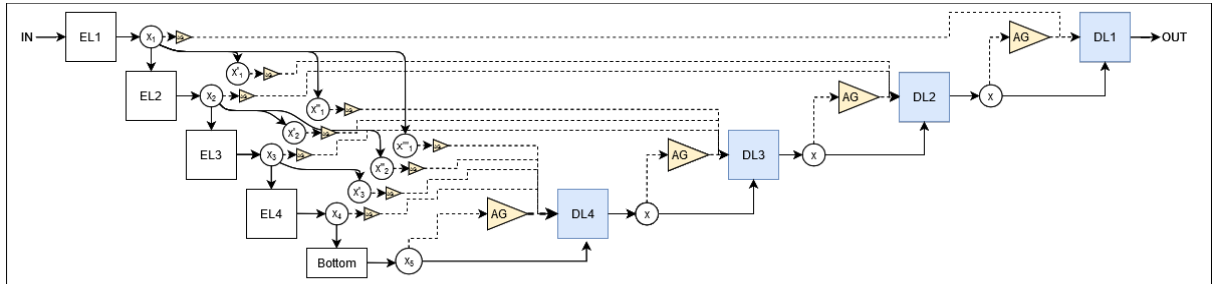

Fig. 5. Architecture of Full Attention U-net

**MODEL TRAINING**

In this section, the training process of 4 networks is elaborated, including dataset preparation, optimization and loss function, result evaluation, hardware configuration, etc. Hyper parameters should be adjusted and finally determined on the basis of the performance of trained models. All the tasks described in this paper are performed in a laptop equipped with an external GPU (GIGABYTE Geforce GTX 1070 mini ITX OC 8G).

*Data Augmentation and Preparation*

The crack photos (see. Fig. 6) revealed by organizers of IPC-SHM 2020 were already classified by a newly developed CNN [22]. For the compromise between hardware limitation for training and the performance of the trained model in prediction, we reshaped 101 original photos with a resolution of 4928×3264 to 386×256 for training and those with a resolution of 4928×3264 to 682×256 for test. Numerous data augmentation strategies have been developed for years [23], each one shows its benefits in different scenarios. In this paper, 3 new photos are generated through flipping a photo around the vertical and/or horizontal directions, so that the training dataset can be enriched 4 times to facilitate model training.

As stated in section INTRODUCTION, experiments are divided into verification and validation. The reasons lay on: a) massive training could be problematic for the limited hardware performance; b) too much interference objects in noise e.g. shadows, marks, weld seams, etc. distribute the training process and afterwards weaken the prediction precision. We prepared a dataset of 30 cell images with a resolution of 512×512 (see examples in Fig. 7) shared by [24] to feed the 4 architectures in training and also test for the competitive study.

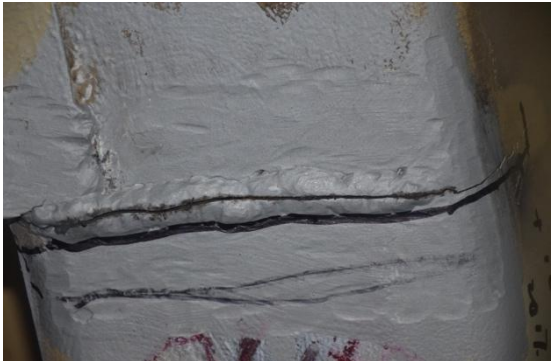 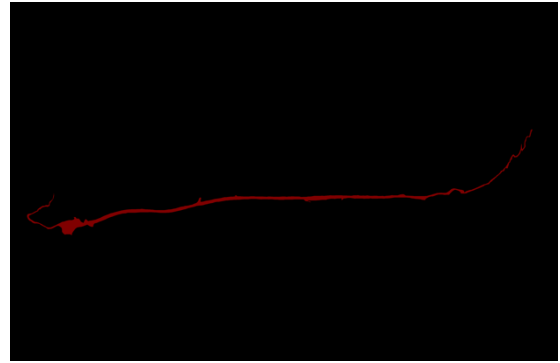

| Image example for training (386×256 pixels) | Label example for training (386×256 pixels) |

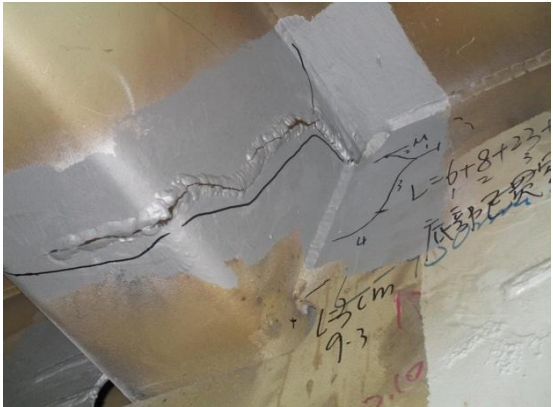 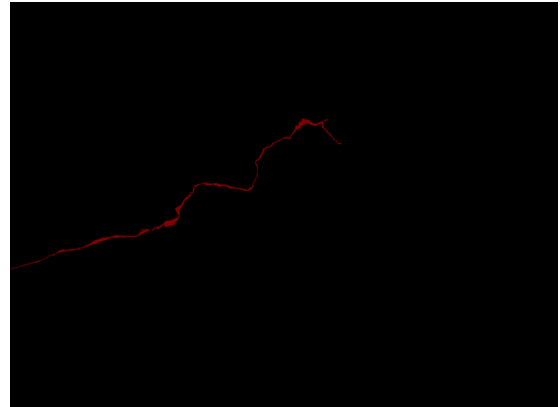

Image example for test (682×256 pixels)　　　　Label example for test (682×256 pixels)

Fig. 6. Examples of crack images and labels for validation

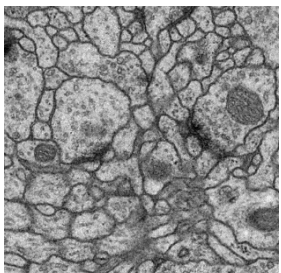 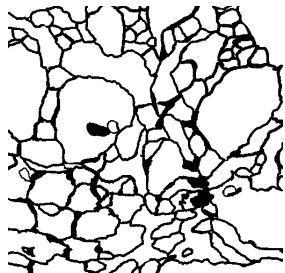 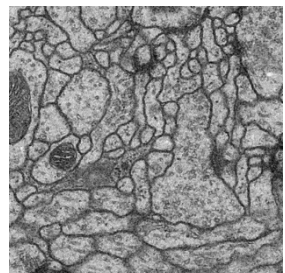 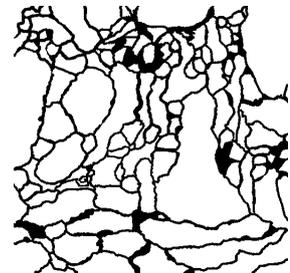

Image example 1　　　Label example.1　　　Image example 2　　　Label example 2
Fig. 7. Examples of cell images and labels for verification [24]

*Optimization*

Ruder [25] summarized the state of the art optimization algorithms. Among the reviewed algorithms in [25], Adaptive Moment Estimation (Adam) [26] is recommended as the eventual best overall choice to address the optimization issue. Adam is a method that computes adaptive learning rates for each parameter. Usually, Adam is interpreted as a stronger version of RMSprop [27] enriched with bias-correction and momentum. Adam is embedded in all the 4 proposed. The parameter values in Adam are defined as 0.9 for the exponential decay rate for the first moment estimates, 0.999 for the exponential decay rate for the second-moment estimates.

*Loss Function*

In a CNN, the degree of deviation between the predicted and actual values of a model in training is represented by loss, which is calculated by the loss function. A larger loss indicates the larger difference between outputs of the model in training and the corresponding labels. On one hand, loss should be minimized as much as possible epoch by epoch in training iteration, thus the change of loss values shows a convergent tendency. On the other hand, the overfitting issue is supposed to be avoided, otherwise the expected results cannot be predicted. Analogous to the review work about optimization, Jadon [2020] investigated loss functions for semantic segmentation nowadays and established an overview, since semantic segmentation is such an active field and model training for this task can benefit from an appropriate, specific loss function through indicating the convergence of loss and affecting backpropagation over training steps. In this paper, we adopt the Binary Cross Entropy (BCE) with Logits Loss. This loss represented in Eq. (1) combines a sigmoid layer and the BCELoss [cite] in one single class.

$$l_n = -w_n[t_n \cdot \log \log \sigma(x_n) + (1 - t_n) \cdot \log \log (1 - \sigma(x_n))]. \quad (1)$$

*Result Evaluation using mIoU*

Intersection over Union (IoU) is an evaluation metric to measure the accuracy of an object detector on a particular dataset. The IoU can be determined via Eq. 2, it indicates the ratio between the area of overlap to the area of union between the perditions and ground truths. Usually, IoU is a metric for a couple of the predictions and the ground truth. For the entire test dataset, mean Intersection over Union (mIoU) is used to check the overall situation of prediction in comparison with the ground truths.

$$IoU = \frac{|A \cap B|}{|A \cup B|} = \frac{|I|}{|U|} \quad (2)$$

**EXPERIMENTS**

*Verification and Competitive Study*

In verification, 4 architectures are trained by the same hyperparameters (epoch = 50, batch size = 2, learning rate = $1e^{-5}$). Under the hyperparameters, all the 4 networks do not reach their highest performance, but the prediction results (see Fig. 8) are already plausible enough for a competitive study. As Fig. 8 shows, in terms of denoising, Attention U-net performs best with most noise, and Advanced Attention U-net contains most noise. Since applying more outputs from each encoding layer improves the feature extraction but also delivers lots of noise, Advanced Attention U-net contains more noise than Attention U-net, however, identifies more distinct cell edges. Black image edges in predictions from Advanced Attention U-net and Full Attention U-net are caused due to the same reason.

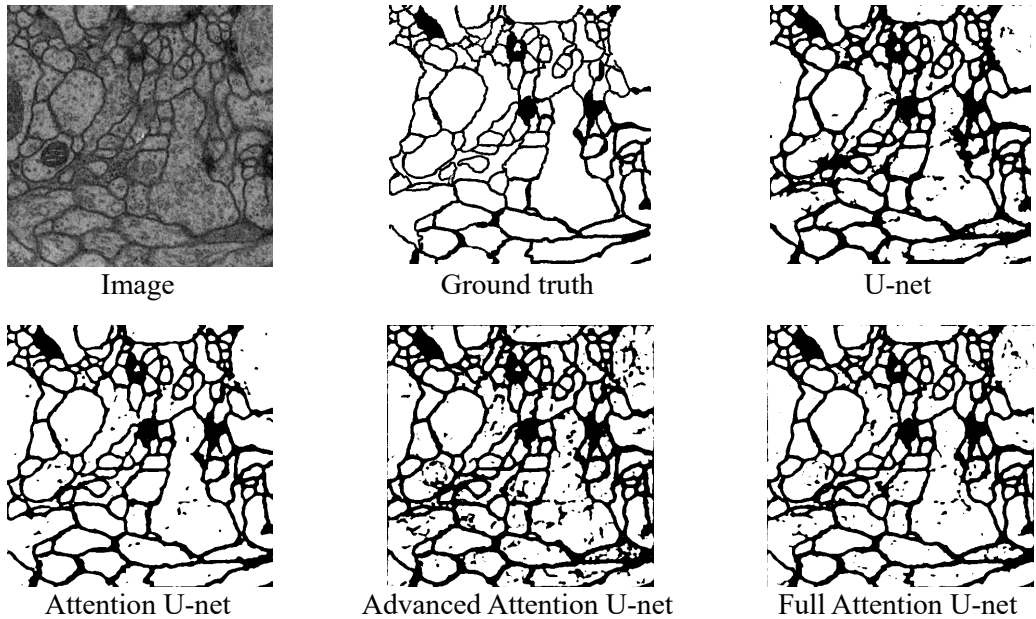

Fig. 8. Result examples of prediction in verification

Though Full Attention U-net and Advanced Attention U-net respectively encompass more noise than Attention U-net and U-net, the mIoU values still show comparable results between these networks. It is, therefore, reasonable to infer that after more training epochs, Full Attention U-net could identify cracks with higher mIoU and less noise than Attention U-net. Hence, we decided to utilize Full Attention U-net works to segment cracks in validation.

Table 1. mIoUs of the prediction results from 4 trained models

| Network | mIoU (%) |
|---|---|
| U-net | 85.59 |
| Attention U-net | 90.85 |
| Advanced Attention U-net | 85.88 |
| Full Attention U-net | 90.02 |

*Validation and Crack Segmentation*

The best score of mIoU we obtained from validation denotes 49.67%. As the prediction results in Fig. 9 display, cracks and some part of noise are segmented in the predated image. The disturbance objects in images mainly misrecognized are the thick line-shaped marks, especially the curves drawn close to cracks, hand-writing numbers and characters.

This research work reveals the short coming of networks using U-net architecture as the backbone. By means of that, the noise, which is similar to objects of interest, cannot be well filtered off, and incur the appearance of noise in the predicted images. This intractable problem needs to be concerned in the future.

IoU of the lower left photo (Prediction 1) demote 48.92%, that of the lower right (Prediction 2) is 46.54%. The former is an overtrained photo, the latter is an incompletely trained one. Visual inspection manifests Prediction 1 contains less noise and also fewer features of the crack than Prediction 2. This causes the area of overlap and that of union, namely numerator and denominator in Eq. (2), decline at the some time. Consequently, the IoU of one image does not fluctuate much in such circumstances.

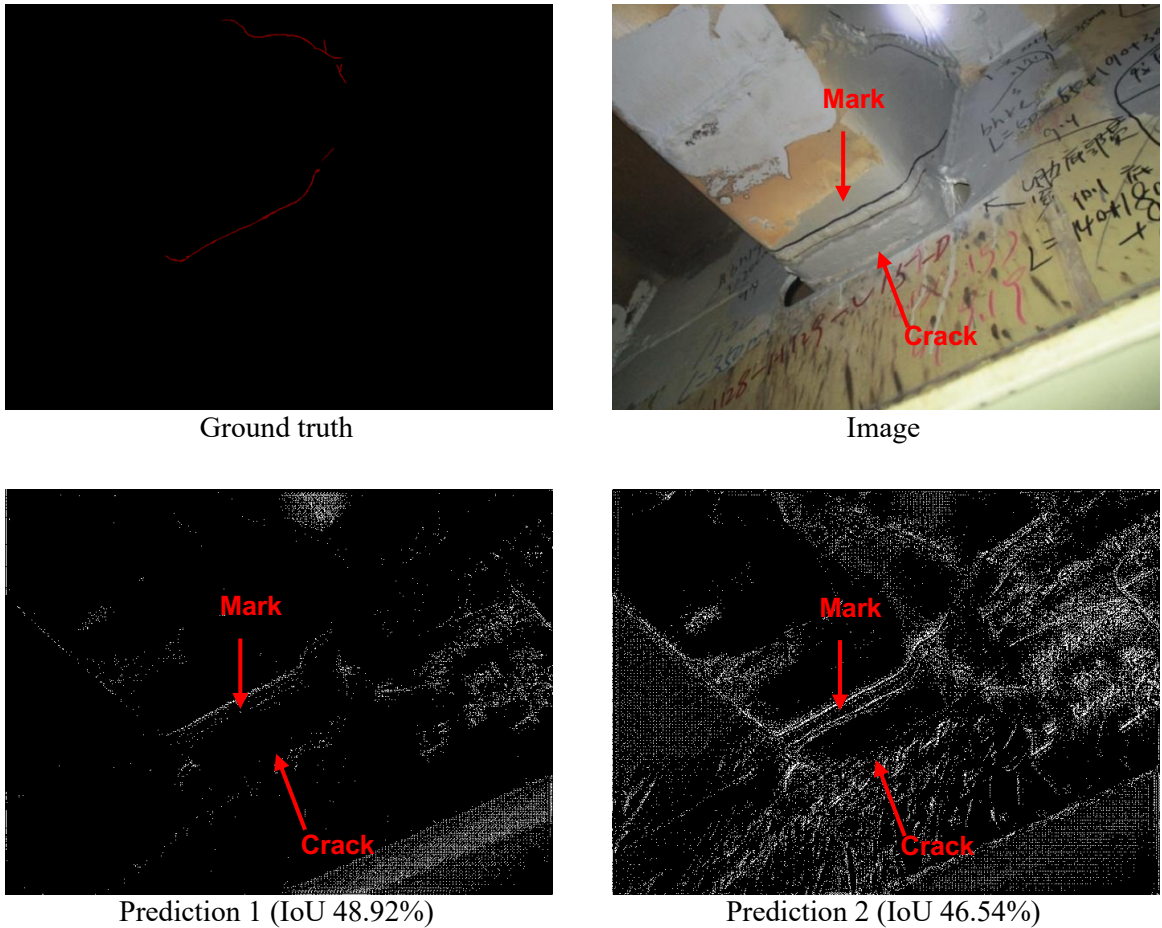

Fig. 9. Result examples of prediction in validation

## DISCUSSION

The experiment results and the proposed architectures are discussed with respect to noise with similar features, impacts from hardware, and metrics for evaluation.

*Noise with similar features*

Our concern was confirmed. The main difference between datasets in verification and validation originate from different types of noise. The noise in cell images are organelles, shapes and sizes of organelles differ from cell edges. On the contrary, interference objects such as marks, weld seams and so on have similar shapes and sizes to cracks. The proposed networks in experiments were not able to distinguish noise from real crack features. If noise are misrecognized as features of interest, it can be unfortunately also focused by the Full Attention Strategy, so that the noise is also detected and segmented as cracks in prediction.

*Impacts from Hardware*

Hardware performance prevents hyperparameters from free tuning. Using a GPU with 8 GB capacity, the batch size is not allowed to exceed 2 in handling images with a resolution of 512×512. Feeding networks on images with higher resolution is also impossible. In images with a low resolution, the continuity of cracks is not well reflected, then influences the quality of segmentation negatively. Besides, time consumption is another issue during the procedures of training and prediction.

*Metrics for Evaluation*

Full Attention Network was trained in validation for times in order to tune hyperparameters. In some scenarios, the mIoU from overfitting prediction results is larger than that from normal prediction results (see Fig. 9). The mIoU is determined as the only matrices and thus might be unable to indicate the prediction quality objectively. Multi-metrics could be more plausible to evaluate the performance of the trained networks.

## CONCLUSION

This paper proposed Full Attention U-net, which is an architecture based on U-net and utilizing attention mechanism and skip connections. 4 networks are tested in cell image segmentation to investigate their performances, and prove the functionality and advantage of Full Attention U-net. Afterwards, Full Attention U-net is hence leveraged to segment crack images for structural health monitoring.

In the future, a better solution should be developed to reduce more noise in the prediction results. Furthermore, more open source datasets can be tested for the comparison with the state-of-the-art results to discover more characters of Full Attention U-net can.

## ACKNOWLEDGEMENTS

In this research work, Mr. Fangzheng Lin constructed the methodology and completed the major part of python codes, Mr. Jiesheng Yang contributed mainly to execution of training and test. This paper is not only for the IPC SHM 2020 but also an extension study of the project BIM-SIS, No. 01-S18017D. The authors would like to acknowledge the support of the Federal Ministry of Education and Research of Germany to the funding of this project.